# Merge-based syntax is mediated by distinct neurocognitive mechanisms: A clustering analysis of comprehension abilities in 84,000 individuals with language deficits across nine languages

Elliot Murphy[1,2], Rohan Venkatesh[3], Edward Khokhlovich[3], Andrey Vyshedskiy[4]

1. Vivian L. Smith Department of Neurosurgery, UTHealth, Texas, USA
2. Texas Institute for Restorative Neurotechnologies, UTHealth, Texas, USA
3. Independent Researcher
4. Boston University, Metropolitan College, Massachusetts, USA

## Abstract

In the modern language sciences, the core computational operation of syntax, 'Merge', is defined as an operation that combines two linguistic units (e.g., 'brown', 'cat') to form a categorized structure ('brown cat', a Noun Phrase). This can then be further combined with additional linguistic units based on this categorial information, respecting non-associativity such that abstract grouping is respected. Some linguists have embraced the view that Merge is an elementary, indivisible operation that emerged in a single evolutionary step. From a neurocognitive standpoint, different mental objects constructed by Merge may be supported by distinct mechanisms: (1) simple command constructions (e.g., "eat apples"); (2) the merging of adjectives and nouns ("red boat"); and (3) the merging of nouns with spatial prepositions ("laptop behind the sofa"). Here, we systematically investigate participants' comprehension of sentences with increasing levels of syntactic complexity. Clustering analyses revealed behavioral evidence for three distinct structural types, which we discuss as potentially emerging at different developmental stages and subject to selective impairment. While a Merge-based syntax may still have emerged suddenly in evolutionary time, responsible for the structured symbolic turn our species took, different cognitive mechanisms seem to underwrite the processing of various types of Merge-based objects.

# Introduction

In modern linguistics, Merge is a proprietary cognitive operation that combines two linguistic units (e.g., 'blue', 'cat') to form a categorized, labeled set ('blue cat', a Noun Phrase), which can then be further combined with additional linguistic units (Chomsky, 1995, 2000). The field of linguistics has largely coalesced around the hypothesis that unbounded Merge is a uniquely human ability, serving as the generative engine underlying the human capacity to generate or communicate an infinite number of expressions (Tanaka et al., 2019). Some researchers, such as Berwick and Chomsky, have noted that due to the 'absolute' nature of Merge (one either has it, or does not have it), no partial or 'proto-Merge' is plausible: "The elementary operation of binary set formation (Merge) appeared in a single step" (Berwick & Chomsky, 2019). From a neurocognitive standpoint, it may be plausible to decompose distinct generic cognitive operations that subserve distinct components of a Merge-based syntax (e.g., the formation of hierarchical objects, the categorization of phrases, the short-term maintenance of categorial identity) (Murphy, 2015).

From the perspective of linguistic formalism, the simplicity of the Merge operation is strikingly elegant. Its concise formulation echoes (intentionally) the parsimony of Hamilton's principle of least action and fundamental physical laws, such as the conservation of energy and the conservation of momentum (Murphy, Holmes, et al., 2024). But from a neurocognitive standpoint, it is becoming increasingly plausible that different levels of Merge-based complexity may be supported by distinct nodes in the extended language network (Murphy, 2020, 2024, 2025; Murphy et al., 2022, 2023; Murphy, Rollo, et al., 2024).

Studies involving large numbers of participants with language deficits offer a powerful approach to investigating the nature of Merge-based syntax. By including tens of thousands of individuals with a range of conditions associated with language impairments, researchers can observe the effect of a wide variety of genetic abnormalities linked to linguistic deficits. Merge interfaces with various conceptual and performance systems, and if some of these exhibit digital/discrete representations but others exhibit analog/graded representations (Jackendoff & Erk, 2025), we may expect to find genetic abnormalities that cause a gradual degradation of syntactic and semantic performance. Conversely, if Merge outputs to a single restricted and homogeneous domain-general interface, its loss should be catastrophic and complete.

Yet, previous studies have shown that neither of these two hypotheses is strongly supported. Two studies of tens of thousands of individuals identified three distinct general levels of what a Merge-based syntax can output: Syntactic, Modifier, and Command (Vyshedskiy, Venkatesh, &

Khokhlovich, 2024; Vyshedskiy, Venkatesh, Khokhlovich, et al., 2024). The Syntactic Phenotype is exhibited by most adults and by children aged four years and older (Vyshedskiy et al., 2025). Individuals with this phenotype can comprehend sentences containing spatial prepositions, reversible word order, verb tenses, possessive pronouns, complex explanations, and elaborate fairytales demanding an array of these construction types (Table 1, *Syntactic Mechanism*). Approximately 2% of adults, as well as children between the ages of three and four, exhibit the Modifier Phenotype. Their ability is limited to combinations of nouns and adjectives (e.g., they can select 'a small green pencil' from a set of pencils, straws, and Lego pieces of different sizes and colors; Table 1, *Modifier Mechanism*). Finally, about 1% of adults, and children aged two to three years, exhibit the Command Phenotype, being restricted to simple commands (e.g., 'eat apple'; Table 1, *Command Mechanism*). Note that these categories delimit overt linguistic performance and major aspects of competence, but may not reflect the full extent of underlying competence.

These findings suggest that while Merge-based syntax can degrade in a seemingly catastrophic manner, it does not vanish entirely. For example, individuals functioning at the intermediate Modifier level lose the ability to comprehend fairytales, spatial prepositions, verb tense, and possessive pronouns, but they retain the capacity to integrate adjectives with nouns, form noun phrases with superlatives, and perform similar compositional tasks. This dovetails with the observation that even in the most severe cases of lesion-based syntactic disruption, some capacity to execute basic hierarchical structure-formation is often preserved (in some format) (Dragoy et al., 2017).

Within this context of open questions, the goals of the present study were: (1) to relate the Command, Modifier, and Syntactic Mechanisms to Merge-based syntax; (2) to analyze a larger cohort of participants than previous work; and (3) to confirm the dissociation among the three Mechanisms within each of the nine languages analyzed individually.

Languages differ significantly in their grammatical structures, word order, word order flexibility, and morphological complexity. For instance, English typically follows a subject–verb–object word order, whereas languages like Japanese and Korean follow a subject–object–verb structure. In English, adjectives are generally pre-nominal—one would say "*the large cat*", not "*the cat large*". In contrast, Romance languages often place adjectives after the nouns they modify. Morphological complexity also varies widely: Russian, for example, features an extensive system of inflectional case endings, in contrast to the relatively simple morphology of English. This greater morphological richness in Russian contributes to its more flexible word order.

If the Command, Modifier, and Syntactic Mechanisms differ across languages, we might be able to associate these grammatical differences with corresponding variations in those mechanisms. Conversely, if no such differences are observed, it would suggest that these three mechanisms—Command, Modifier, and Syntactic—may be universal across languages.

## Methods

### Study Participants

Participants were children and adolescents using a language therapy app that was made freely available at all major app stores in September 2015 (Dunn, Elgart, Lokshina, Faisman, Khokhlovich, et al., 2017b, 2017a; Dunn, Elgart, Lokshina, Faisman, Waslick, et al., 2017; Vyshedskiy et al., 2020; Vyshedskiy & Dunn, 2015). The app provides various structured language comprehension therapy exercises and is primarily used by caregivers of children with language impairments. Most of the caregivers are presumed to be parents. Once the app was downloaded, caregivers were asked to register and to provide demographic details, including the child's diagnosis and age. Caregivers consented to pseudonymized data analysis and completed a 133-item questionnaire (77-item Autism Treatment Evaluation Checklist (ATEC) (Rimland & Edelson, 1999), Supplementary Tables 1-4; 20-item Mental Synthesis Evaluation Checklist (MSEC) (Braverman et al., 2018), Supplementary Table 5; 10-item screen time checklist (Fridberg et al., 2021); 25-item diet checklist (Acosta et al., 2023); and 1-item parent education survey) approximately every three months.

All fifteen available *language comprehension* items from the 133-item questionnaire were included in the cluster analysis as in previously published articles (Vyshedskiy, Venkatesh, & Khokhlovich, 2024; Vyshedskiy, Venkatesh, Khokhlovich, et al., 2024) (Table 1). Answer choices were as follows: very true (0 points), somewhat true (1 point), and not true (2 points). A lower score indicates better language comprehension ability.

The inclusion criteria for this study remained consistent with those of previous studies (Vyshedskiy, Venkatesh, & Khokhlovich, 2024; Vyshedskiy, Venkatesh, Khokhlovich, et al., 2024): absence of seizures (which commonly result in intermittent, unstable language comprehension deficits (Forman et al., 2022)), absence of serious and moderate sleep problems (which are also associated with intermittent, unstable language comprehension deficits (Levin et al., 2022)), age range of 4 to 22 years (the lower age cutoff was chosen to ensure that participants were exposed to complete set of sentence structures listed in Table 1 (Arnold & Vyshedskiy, 2022); the upper age cutoff was chosen to avoid analysis of participants who may be linguistically declining due to aging). Previous studies were limited to individuals diagnosed with Autism Spectrum Disorder (ASD) (Vyshedskiy, Venkatesh, & Khokhlovich, 2024) and individuals with fluid speech (Vyshedskiy, Venkatesh, Khokhlovich, et al., 2024). This study included all participants who submitted their assessments through the app, speaking one of the nine languages that the app is available in: English, Spanish, Portuguese, Italian, Russian, Chinese, French, German, and Korean. Table 2 reports participants' demographics in each language group as communicated by caregivers. Males outnumber females by approximately four to one, reflecting the predominance of autism among participants and its known male-to-female prevalence ratio.

Table 3 reports participants' diagnoses as communicated by caregivers. Autism level (mild/Level 1, moderate/Level 2, or severe/Level 3) was reported by caregivers. Pervasive Developmental Disorder and Asperger Syndrome were combined with mild autism for analysis as recommended by DSM-5 (American Psychiatric Association, 2013). A good reliability of such parent-reported diagnosis has been previously demonstrated (Jagadeesan et al., 2022).

When caregivers have completed several evaluations, the last evaluation was used for analysis as in previous studies (Vyshedskiy, Venkatesh, & Khokhlovich, 2024; Vyshedskiy, Venkatesh, Khokhlovich, et al., 2024). Thus, the study included a total of 84,099 participants, the average age was 6.5 ± 2.7 years (range of 4 to 21.9 years), 74.7% participants were males. The education level of participants' parents was the following: 90.9% with at least a high school diploma, 68.6% with at least college education, 35.8% with at least a master's, and 5.6% with a doctorate. All caregivers consented to pseudonymized data analysis and publication of results. The study was conducted in compliance with the Declaration of Helsinki (World Medical Association, 2013). Using the Department of Health and Human Services regulations found at 45 CFR 46.101(b)(4), the Biomedical Research Alliance of New York (BRANY) LLC Institutional Review Board (IRB) determined that this research project is exempt from IRB oversight (IRB File # 22-12-205-1120). The data was accessed on July 9, 2025.

**Statistics and Reproducibility**

Unsupervised Hierarchical Cluster Analysis (UHCA) was performed using Ward's agglomeration method with a Euclidean distance metric. The clustering analysis was data-driven without any design or hypothesis. A two-dimensional heatmap was generated using the "pheatmap" package of R, freely available language for statistical computing (R Foundation for Statistical Computing, 2021). Code and data can be downloaded (doi.org/10.17605/OSF.IO/2QK5B).

## Results

### Clustering analysis of 15 language comprehension abilities

Caregivers assessed 15 language comprehension abilities (Table 1). To examine patterns of co-occurrence among these abilities, we applied unsupervised hierarchical cluster analysis—a data-driven method that groups items based on their similarity. This technique produces tree-like diagrams, called *dendrograms*, which visually represent the hierarchical relationships between clusters of items. Abilities that frequently co-occur are positioned closely together, while those that co-occur less often appear farther apart.

Figure 1A depicts the dendrogram generated from the analysis of English-speaking participants. The height of the branches indicates the distance between clusters. A larger distance corresponds to greater dissimilarity between the clusters. Previous studies identified three clusters stable across different evaluation methods, age groups, time points, genders, and parental education (Vyshedskiy, Venkatesh, & Khokhlovich, 2024; Vyshedskiy, Venkatesh, Khokhlovich, et al., 2024). The first cluster included *knowing the name, responding to 'No' or 'Stop', responding to praise*, and *following some commands* (items 1 to 4 in Table 1) and was termed the Command Mechanism. The second cluster included *understanding color and size modifiers, several modifiers in a sentence, size superlatives*, and *numbers* (items 5 to 8 in Table 1) and was termed the Modifier Mechanism. The third cluster included *understanding of spatial prepositions, verb tenses, flexible syntax, possessive pronouns, explanations about people and situations, simple stories*, and *elaborate fairytales* (items 9 to 15 in Table 1) and was termed the Syntactic Mechanism. The analysis of English-speaking participants in Figure 1A identified the same three clusters with inter-cluster distances that were significantly larger than the distances between subclusters. Principal component analysis (PCA) (Figure 1B) also showed a clear separation between the three clusters: Command, Modifier and Syntactic.

As a control we calculated unsupervised hierarchical cluster analysis and PCA of the 15 comprehension abilities along with the "*hyperactivity*" (Figures S1), "*bed-wetting*" (Figure S2), and "*demands sameness*" (Figure S3) items. These items are not related to language and therefore should cluster into their own group. As expected, both unsupervised hierarchical cluster analysis and PCA clustered these items into their own group at a significant distance from the Command, Modifier, and Syntactic clusters, validating both clustering techniques.

## Clustering analysis across spoken languages

Clustering analysis was conducted in all language groups with 400 or more participants. The three-cluster solution was consistent across all language groups explored: English, Spanish, Portuguese, Italian, Russian, Chinese, French, German, and Korean (Figures 1-9). In all spoken languages, unsupervised hierarchical cluster analysis sorted the 15 comprehension abilities into congruent three clusters (Command, Modifier, and Syntactic) and PCA showed a clear separation between the three clusters. Some language groups, such as Russian (Figure 5B), demonstrated a greater separation between clusters in PCA.

These findings suggest that the three-cluster solution is not a result of differential cultural upbringing but rather a potentially general cognitive phenomenon constrained, consistent across spoken languages.

## Language comprehension phenotypes in participants

Previous studies have employed unsupervised hierarchical cluster analysis to identify distinct language comprehension phenotypes of participants (Vyshedskiy, Venkatesh, & Khokhlovich, 2024; Vyshedskiy, Venkatesh, Khokhlovich, et al., 2024). The principles underlying participant clustering are identical to those used for clustering abilities: participants with similar patterns of abilities are automatically organized into hierarchical dendrograms. Previous studies identified three distinct phenotypes: 1) Command Phenotype–participants who acquired only the Command Mechanism; 2) Modifier Phenotype–participants who acquired both the Command and Modifier Mechanisms; and 3) Syntactic Phenotype–participants who acquired the Command, Modifier, and Syntactic Mechanisms.

The close correspondence between comprehension mechanisms and the resulting phenotypes is noteworthy. While various combinations of the three mechanisms are theoretically possible, such combinations were not observed empirically. For example, a hypothetical phenotype combining the Command and Syntactic Mechanisms (but not the Modifier Mechanism) could exist in theory. Another possibility would be a phenotype lacking all three mechanisms. However, these configurations did not emerge from the data. This absence suggests that the 'morphospace' of comprehension phenotypes is constrained by cognitive faculties.

To investigate whether these constraints are consistent across different languages, this study conducted the unsupervised hierarchical cluster analysis of participants separately within each language group. The results are presented in Figures 10-18, which relate participant clusters to comprehension mechanisms. The three mechanism clusters (Command, Modifier, and Syntactic) are shown as rows (the dendrogram from Figure 1A representing comprehension mechanisms is shown vertically on the left in Figure 10) and the 27,187 English-speaking participants are shown as columns (the dendrogram representing participants is shown horizontally on the top). *Blue* indicates the presence of a linguistic ability (parent's response=*very true*); *white* indicates an intermittent presence of a linguistic ability (parent's response=*somewhat true*); and *red* indicates the complete lack of a linguistic ability (parent's response=*not true*).

In the heatmap of English-speakers (Figure 10), the middle cluster of participants (marked "Syntactic Phenotype") shows the predominant blue color (representing good skills) across all abilities indicating that these participants acquired the Command, Modifier, and Syntactic Mechanisms. The leftmost cluster of participants (marked "Command Phenotype") shows the predominant blue color only among the Command Mechanism items and red colors across Syntactic and Modifier Mechanisms items, indicating that these individuals only acquired the Command Mechanism. The rightmost cluster of participants (marked "Modifier Phenotype") shows the predominant blue color only across Command and Modifier Mechanisms items and white to red colors across Syntactic Mechanism items, indicating that these individuals acquired the Command and Modifier Mechanisms.

This pattern was reproduced across all language groups (Figures 10-18). Participants acquired either: 1) the Command Mechanism alone (marked as the Command Phenotype), or 2) both the Command and Modifier Mechanisms (marked as the Modifier Phenotype), or 3) the Command, Modifier, and Syntactic Mechanisms (marked as the Syntactic Phenotype).

## Discussion

We conducted a clustering analysis to examine the co-occurrence of fifteen language comprehension abilities in 84,099 individuals who spoke English, Spanish, Portuguese, Italian, Russian, Chinese, French, German, or Korean. The three identified clusters were identical between languages and congruent to those found in previous analyses (Vyshedskiy, Venkatesh, & Khokhlovich, 2024; Vyshedskiy, Venkatesh, Khokhlovich, et al., 2024). Crucially, the clustering analysis in all studies was devoid of any design or hypothesis, as both unsupervised hierarchical clustering analysis and principal component analysis (PCA) were entirely driven by the data. The outcome of our clustering analyses is a set of three coherent, discrete language ability clusters that appear to revolve around similar linguistic deficits concerning modifiers and complex syntactic operations—not a mixed amalgam of different patterns that would be expected if language abilities were mediated by many unrelated mechanisms.

We note here some limitations of our work: A 133-item survey completed repeatedly by motivated parents is invaluable, but still prone to optimism, fatigue, and socioeconomic skew. Future work could strengthen the validity of findings by quantifying inter-rater reliability (e.g., by having both parents independently complete the survey). While one study has demonstrated a strong correlation between the parent-reported survey used in this study and a clinicians-administered Language Phenotype Assessment ($r$ = 0.78, $p$ < 0.0001) (Vyshedskiy et al., 2025), further validation could involve comparing a subsample of parent responses with results from standardized clinician-administered instruments (e.g., PLS-5 (Zimmerman et al., 2011), Token Test (A. De Renzi & Vignolo, 1962; E. De Renzi & Faglioni, 1978), CELF-5 (Wiig et al., 2013), TROG (Bishop, 2009)).

Our enrollment protocol also filters for relatively "tech-savvy", intervention-seeking parents and may under-represent low-SES households. A comparison of census-matched demographics would help establish external validity. In addition, the disorders reported in our study differ in etiology and linguistic phenotype; since the same caregiver questionnaire supplies both phenotype and explanatory variable, latent correlations may inflate cluster separability.

The replication of the three-cluster solution across English, Spanish, Portuguese, Italian, Russian, Chinese, French, German, and Korean likely points to language-independent constraints. Notable, the languages examined in this study vary widely in morphological complexity, word order and word order flexibility. For example, Russian has a much richer morphological system and greater word order flexibility than English; while Korean follows subject–object–verb structure, which differs from the subject–verb–object order typical of Indo-European languages. Despite these structural differences, all nine languages consistently revealed clear distinctions among the three linguistic mechanisms—Command, Modifier, and Syntactic—indicating that these distinctions may be universal.

Some may argue here that the three clusters reflect emergent probabilistic constructions rather than discrete cognitive mechanisms. However, the sharp boundaries we observe (especially the near-absence of Modifier-without-Command or Syntactic-without-Modifier profiles) are difficult to reconcile with a purely continuous competence model. Future work could operationalize the predictions of usage-based linguistic theories to more carefully explore these topics.

The present behavioral gradient (Command → Modifier → Syntactic) demonstrates that a single explanatory construct, Merge, can be developmentally unpacked into sub-routines mastered at different developmental stages: the Command Mechanism by 1.6 years of age, the Modifier mechanism by 3.0 years of age, and the Syntactic Mechanism by 3.7 years of age (Vyshedskiy et al., 2025). From an evolutionary perspective, this aligns with a "saltation-plus-scaffolding" model: an initial binary set-forming capacity may have arisen abruptly (Berwick & Chomsky, 2019), but its efficient deployment in real-time cognition and communication required incremental recruitment of domain-general resources such as working memory, attention (Murphy, 2019, 2024) and articulate speech (Vyshedskiy, 2022). It is possible that the layered neurocognitive architecture that supports modern human syntax may provide some basis for the behavioral results we document here.

Isolating behavioral dynamics of "Command", "Modifier", and "Syntactic" mechanisms might allow us to refine long-standing psycholinguistic debates about the grain-size of syntactic representations accessed during real-time comprehension. For example, classic garden-path effects show that comprehenders incrementally commit to local syntactic analyses that sometimes require costly reanalysis when later input forces revision. If the Modifier mechanism licenses phrase-internal operations such as adjective-noun union, while the Command mechanism governs clausal argument structure, then garden-path costs should be sharply magnified whenever disambiguation pivots on Command-level information (e.g., NP-attachment vs. VP-attachment ambiguities). Conversely, ambiguities resolvable within the Modifier mechanism (e.g., prenominal adjective stacking) should yield milder slow-downs.

Our results suggest that the conceptual and performance systems that access Merge-based syntax are fractionated behaviorally into distinct mechanistic groups. We hope that these results play some role in addressing a long-standing gap between formal linguistic theory and large-scale behavioral phenotyping. Our reported sample is an order of magnitude larger than typical language-impairment studies, improving power to detect stable substructures, but many further questions remain concerning the granularity of the documented sets of language deficits.

## Funding

This research received no external funding.

## Acknowledgements

We wish to thank all participants' caregivers who found time to complete children's assessments. The language therapy app used to collect the data presented in this manuscript was made

|  |  | Language comprehension items (verbatim) | Abbreviations used in dendrograms |
|---|---|---|---|
| Command Mechanism | 1 | Knows own name | Knows Name |
|  | 2 | Responds to 'No' or 'Stop' | No and Stop |
|  | 3 | Responds to praise | Resp. to Praise |
|  | 4 | Can follow some commands | Commands |
| Modifier Mechanism | 5 | Understands some simple modifiers (i.e., green apple vs. red apple or big apple vs. small apple) | Color or Size / Modifiers |
|  | 6 | Understands several modifiers in a sentence (i.e., small green apple) | Two Modifiers |
|  | 7 | Understands size (can select the largest/smallest object out of a collection of objects) | Size Superlatives |
|  | 8 | Understands NUMBERS (i.e., two apples vs. three apples) | Numbers |
| Syntactic Mechanism | 9 | Understands spatial prepositions (i.e., put the apple ON TOP of the box vs. INSIDE the box vs. BEHIND the box) | Sp. Prepositions |
|  | 10 | Understands verb tenses (i.e., I will eat an apple vs. I ate an apple) | Verb Tenses |
|  | 11 | Understands simple stories that are read aloud | Simple Stories |
|  | 12 | Understands elaborate fairytales that are read aloud (i.e., stories describing FANTASY creatures) | Elab. Fairytales |
|  | 13 | Understands possessive pronouns (i.e., your apple vs. her apple) | Poss. Pronouns |
|  | 14 | Understands the change in meaning when the order of words is changed (i.e., understands the difference between 'a cat ate a mouse' vs. 'a mouse ate a cat') | Flexible Syntax |
|  | 15 | Understands explanations about people, objects or situations beyond the immediate surroundings (e.g., "Mom is walking the dog," "The snow has turned to water"). | Explanations |

**Table 1. Three language comprehension mechanisms—Syntactic, Modifier, and Command—have been identified in previous studies** (Vyshedskiy, Venkatesh, & Khokhlovich, 2024; Vyshedskiy, Venkatesh, Khokhlovich, et al., 2024). When one mechanism is acquired, the entire range of associated comprehension abilities is also gained. The Command-level abilities (Items 1 to 4) are acquired first. The Modifier-level abilities (Items 5 to 8) are attained next. The Syntactic-level abilities (Items 9 to 15) are acquired last. The language comprehension items are presented exactly as surveyed with parents in both this and earlier studies. Response options

were: very true, somewhat true, and not true. Items 1 to 3 were assessed as part of the Expressive Language ATEC (Rimland & Edelson, 1999) subscale 1; the rest of items were a part of the MSEC subscale (Braverman et al., 2018).

|  | Number of Participants | Percent of Total | Age, Mean(SD) | Percent Males |
|---|---|---|---|---|
| **English** | 27187 | 32.3 | 6.7(2.9) | 75.8 |
| **Spanish** | 33488 | 39.8 | 6.1(2.2) | 68.3 |
| **Portuguese** | 7504 | 8.9 | 6.3(2.5) | 76.3 |
| **Italian** | 6484 | 7.7 | 7.6 (3.2) | 78.3 |
| **Russian** | 4778 | 5.7 | 6.8(2.5) | 77.9 |
| **Chinese** | 2217 | 2.6 | 6.1(2.0) | 79.2 |
| **French** | 1060 | 1.3 | 7.2(3.2) | 72.5 |
| **German** | 927 | 1.1 | 7.2(3.2) | 69.9 |
| **Korean** | 454 | 0.5 | 6.6(2.7) | 74.2 |
| **Total** | **84099** | **100** | **6.5(2.7)** | **74.7** |

**Table 2. Participants' demographics.**

|  | Number of Participants | Percent of Total | Age, Mean(SD) | Percent Males |
|---|---|---|---|---|
| **Mild Autism Spectrum Disorder (ASD)** | 29292 | 34.8 | 6.2(2.4) | 75.8 |
| **Moderate ASD** | 19094 | 22.7 | 6.9(2.8) | 79.5 |
| **Severe ASD** | 10839 | 12.9 | 7.6(3.3) | 80.0 |
| **Not-diagnosed** | 9869 | 11.7 | 5.7(1.6) | 54.2 |
| **Specific Language Impairment** | 4884 | 5.8 | 6.0(2.2) | 69.2 |
| **Mild Language Delay** | 3973 | 4.7 | 5.4(1.7) | 67.4 |
| **ADHD** | 1630 | 1.9 | 6.4(2.2) | 72.8 |
| **Down Syndrome** | 1427 | 1.7 | 8.5(3.5) | 60.4 |
| **Other Genetic Disorder** | 1104 | 1.3 | 8.0(3.5) | 57.7 |
| **Social Communication Disorder** | 794 | 0.9 | 6.3(2.3) | 69.1 |
| **Sensory Processing Disorder** | 727 | 0.9 | 6.8(2.7) | 70.2 |
| **Apraxia** | 466 | 0.6 | 6.8(2.9) | 64.6 |
| **Total** | **84099** | **100** | **6.5(2.7)** | **74.7** |

**Table 3. Participants' diagnoses.**

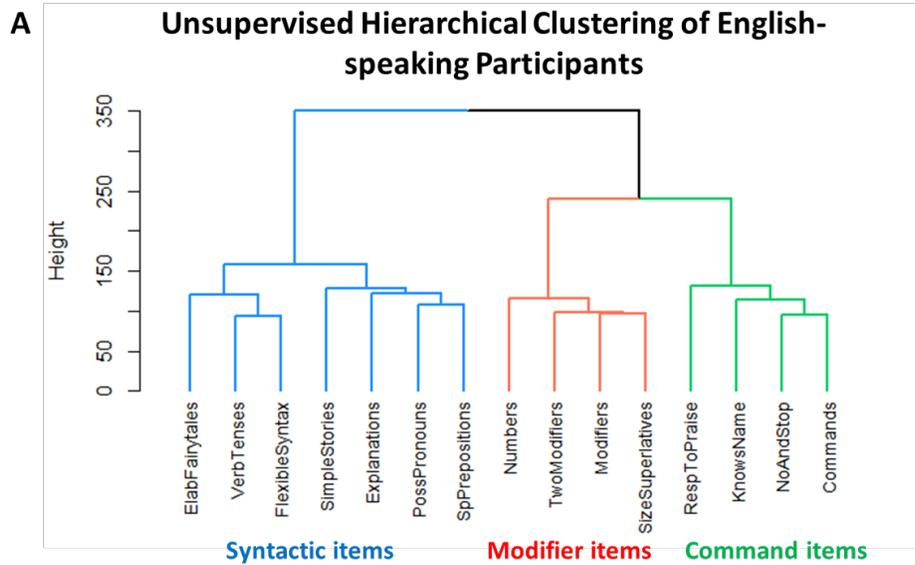

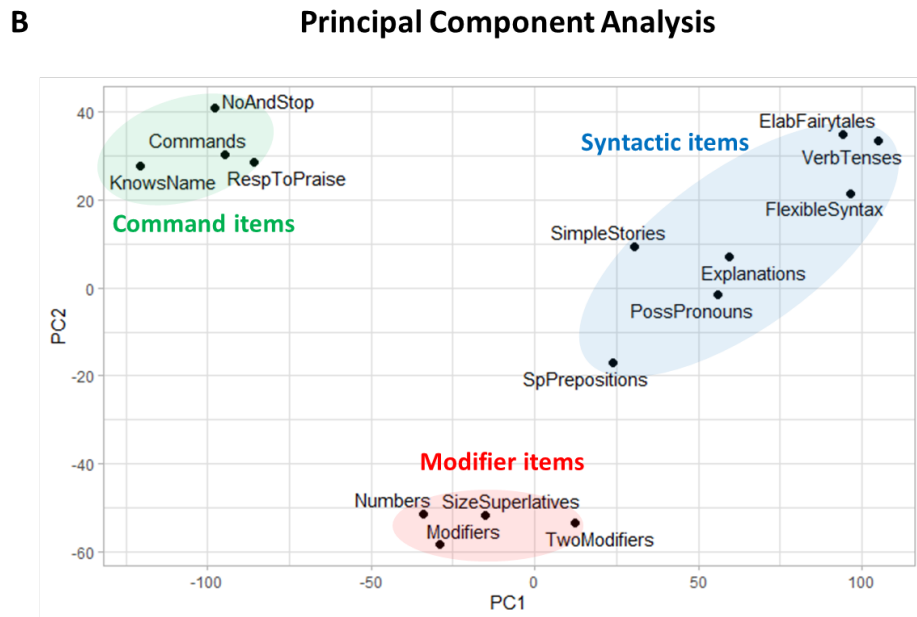

**Figure 1. Clustering analysis of 15 comprehension items in English-speaking participants.** (A) A dendrogram representing the unsupervised hierarchical clustering analysis (UHCA) of language comprehension abilities. (B) Principal component analysis (PCA) plot, where ovals highlight clusters identified by UHCA. The PCA reveals a distinct separation among Command, Modifier and Syntactic Mechanisms. Principal component 1 accounts for 47% of the variance in the data. Principal component 2 accounts for 11.1% of the variance in the data.

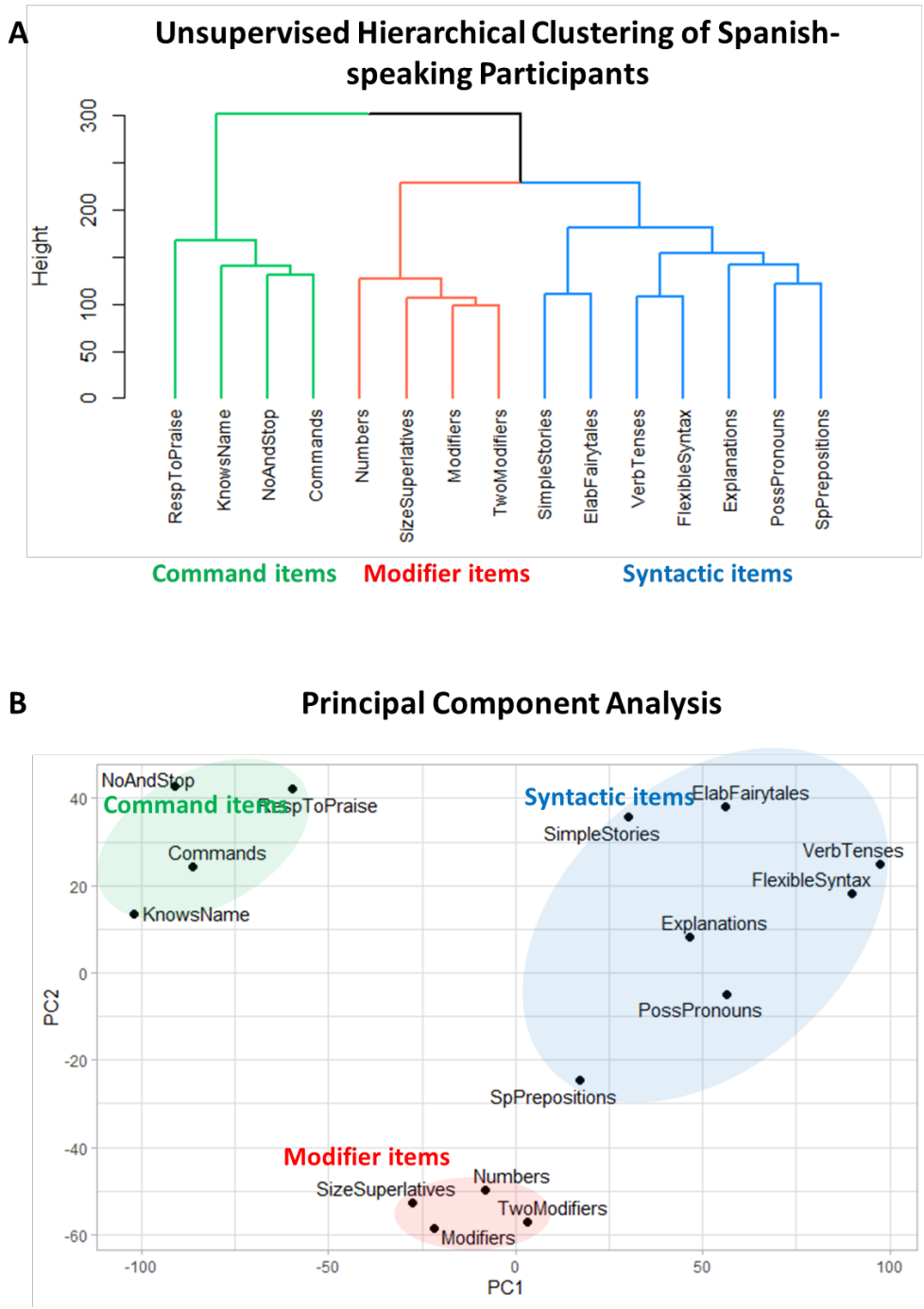

**Figure 2. Clustering analysis of 15 language comprehension items in Spanish-speaking participants.** (A) A dendrogram representing the unsupervised hierarchical clustering analysis of language comprehension abilities. (B) Principal component analysis (PCA) plot, where ovals highlight clusters identified by UHCA. The PCA reveals a distinct separation among Command, Modifier and Syntactic Mechanisms. Principal component 1 accounts for 32.4% of the variance in the data. Principal component 2 accounts for 11.5% of the variance in the data.

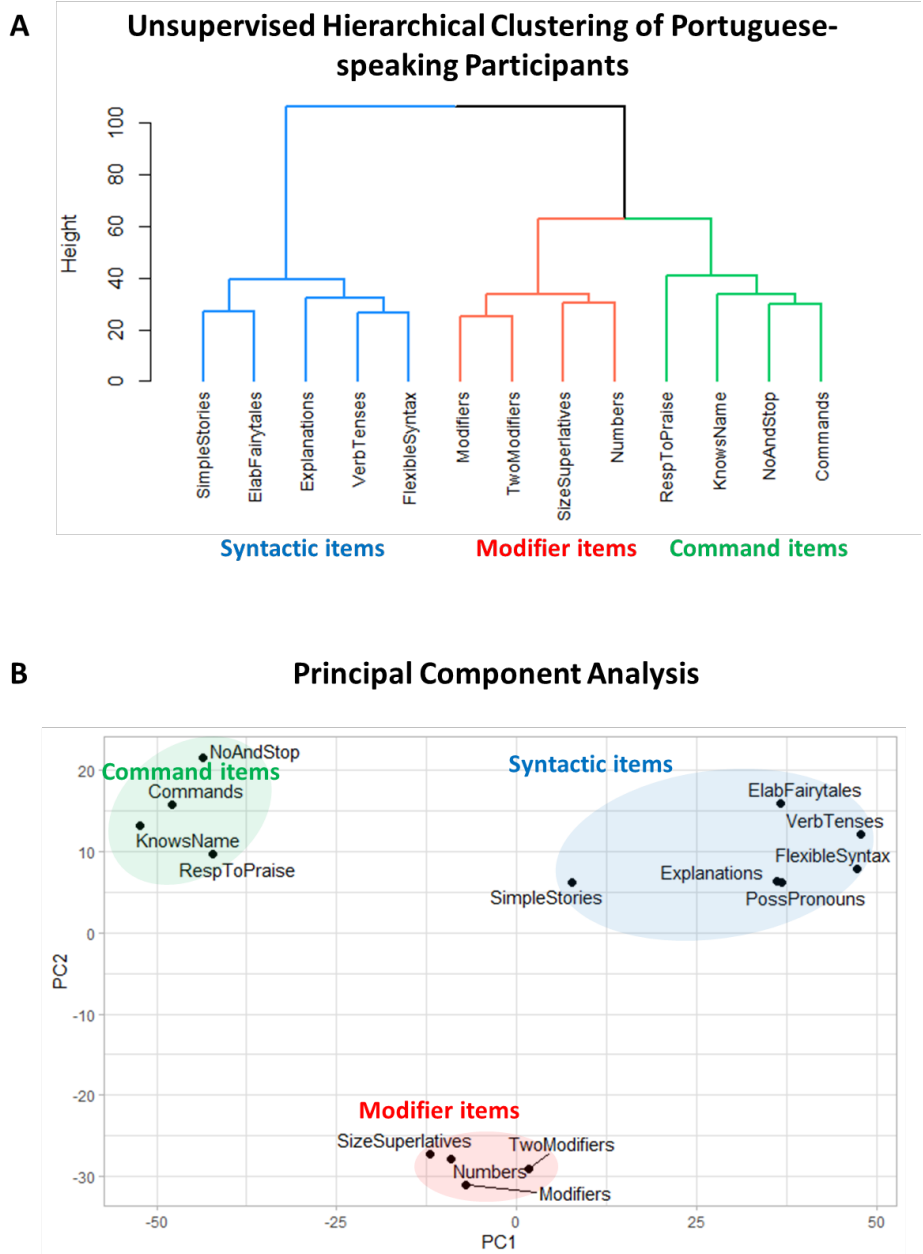

**Figure 3. Clustering analysis of 14 language comprehension items in Portuguese-speaking participants.** One item ("spatial prepositions") was translated incorrectly and was therefore excluded from analysis. (A) A dendrogram representing the hierarchical clustering of language comprehension abilities. (B) Principal component analysis (PCA) plot, where ovals highlight clusters identified by UHCA. The PCA reveals a distinct separation among Command, Modifier and Syntactic Mechanisms. Principal component 1 accounts for 40.4% of the variance in the data. Principal component 2 accounts for 11.3% of the variance in the data.

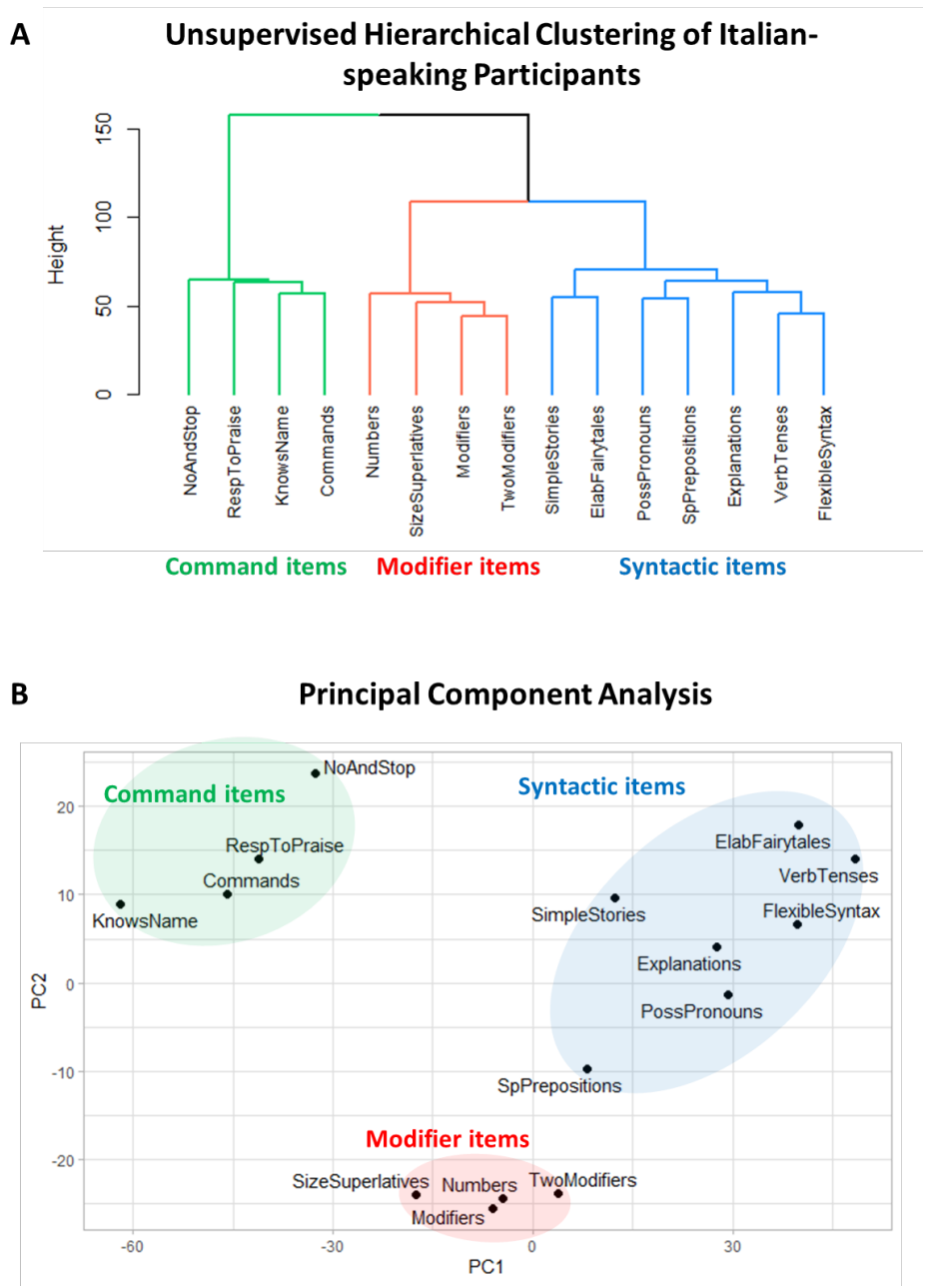

**Figure 4. Clustering analysis of 15 language comprehension items in Italian-speakers.** (A) A dendrogram representing the hierarchical clustering of language comprehension abilities. (B) Principal component analysis (PCA) plot, where ovals highlight clusters identified by UHCA. The PCA reveals a distinct separation among Command, Modifier and Syntactic Mechanisms. Principal component 1 accounts for 43% of the variance in the data. Principal component 2 accounts for 10.8% of the variance in the data.

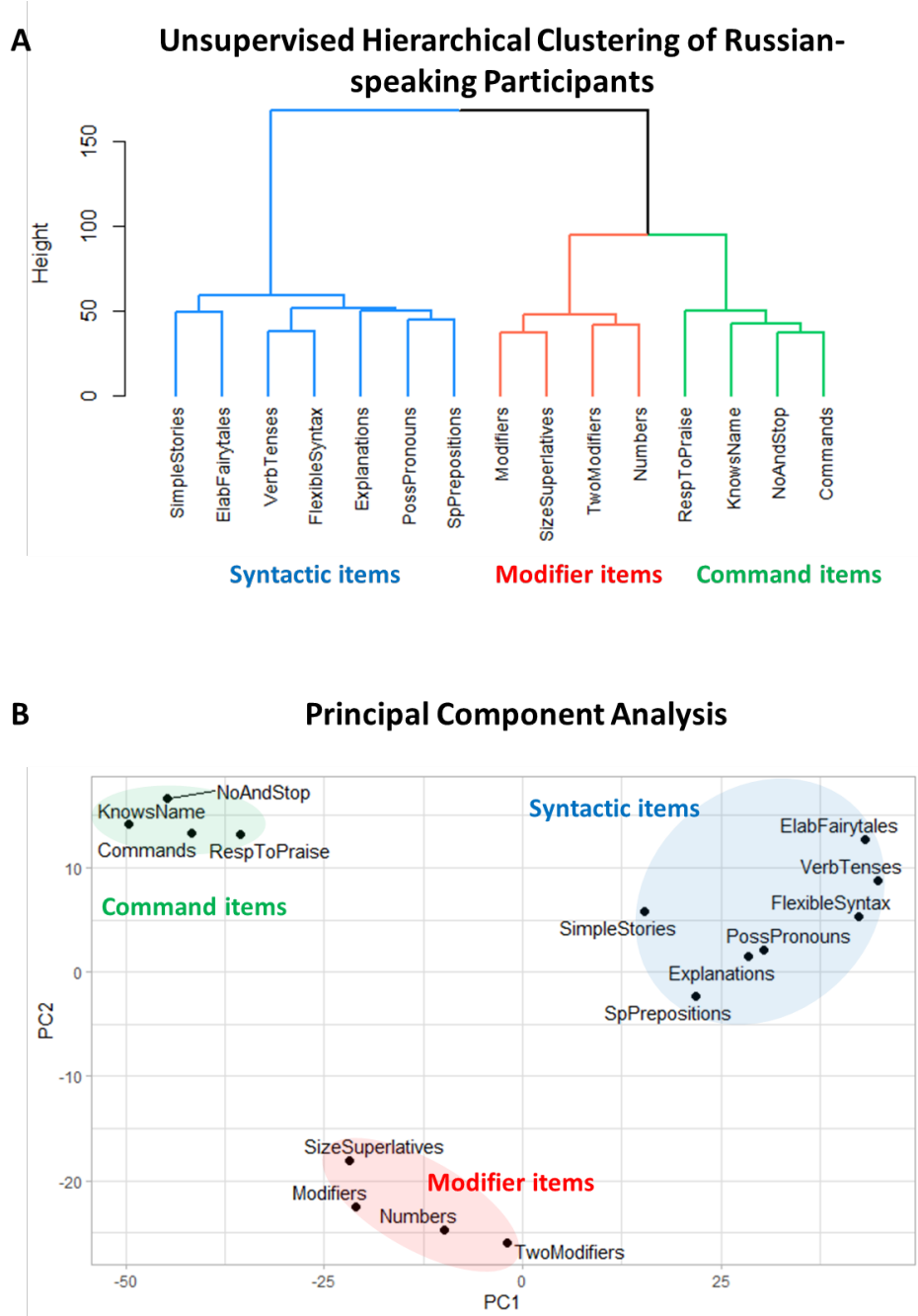

**Figure 5. Clustering analysis of 15 language comprehension items in Russian-speaking participants.** (A) A dendrogram representing the hierarchical clustering of language comprehension abilities. (B) Principal component analysis (PCA) plot, where ovals highlight clusters identified by UHCA. The PCA reveals a distinct separation among Command, Modifier and Syntactic Mechanisms. Principal component 1 accounts for 52.7% of the variance in the data. Principal component 2 accounts for 10.3% of the variance in the data.

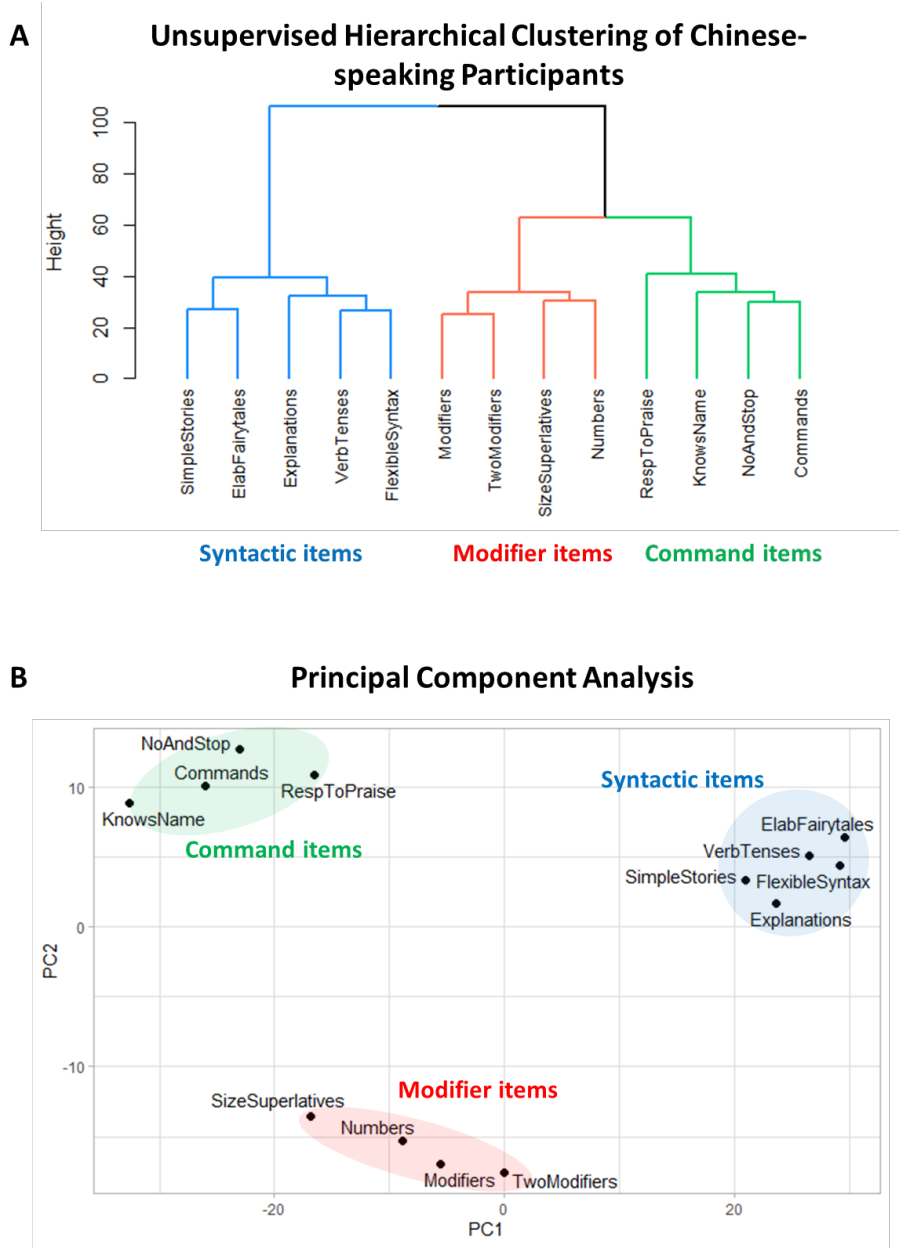

**Figure 6. Clustering analysis of 13 language comprehension items in Chinese-speaking participants.** Two items ("spatial prepositions" and "possessive pronouns") were translated incorrectly and were therefore excluded from analysis. (A) A dendrogram representing the hierarchical clustering of language comprehension abilities. (B) Principal component analysis (PCA) plot, where ovals highlight clusters identified by UHCA. The PCA reveals a distinct separation among Command, Modifier and Syntactic Mechanisms. Principal component 1 accounts for 49.7% of the variance in the data. Principal component 2 accounts for 12.3% of the variance in the data.

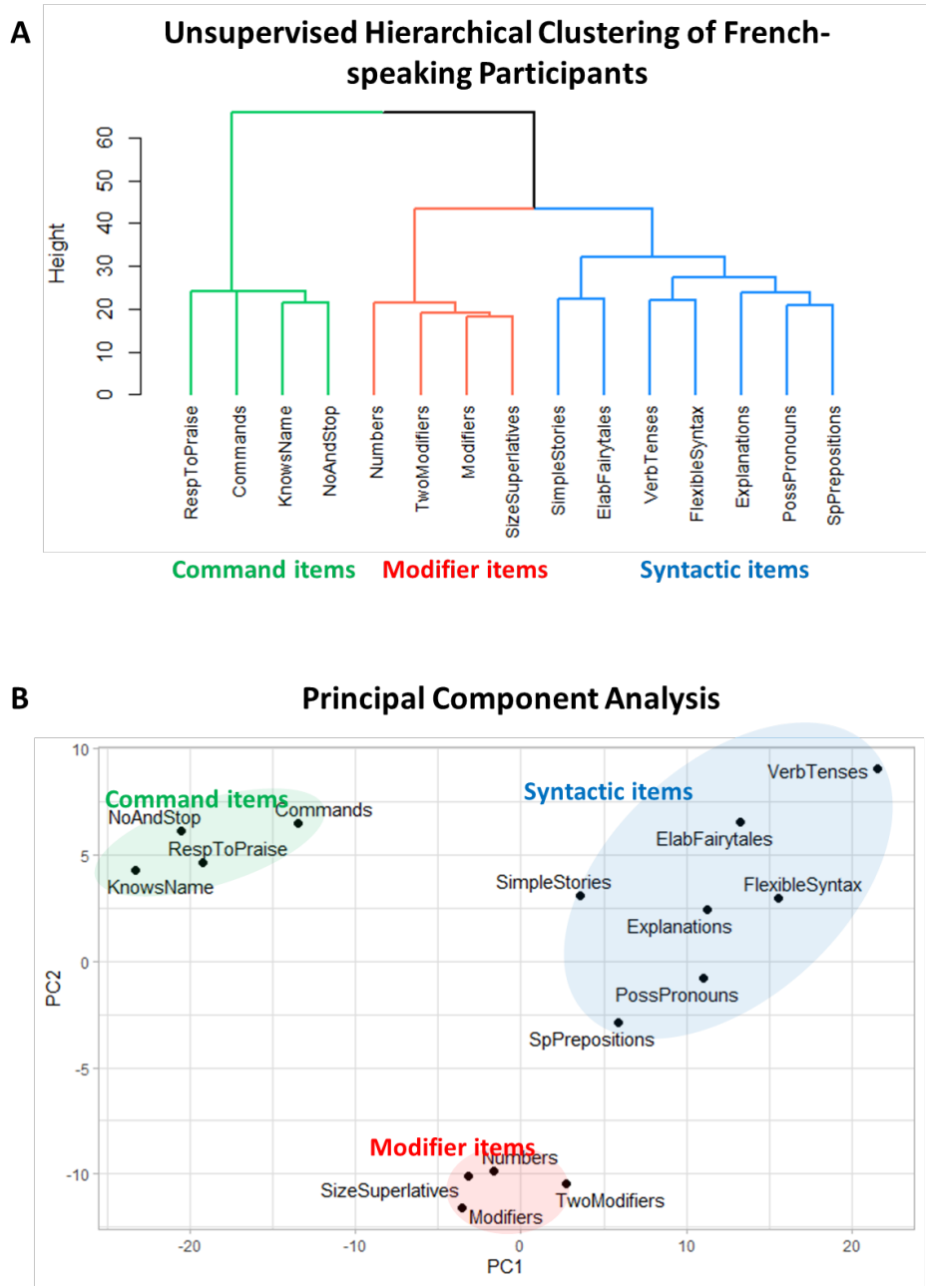

**Figure 7. Clustering analysis of 15 language comprehension items in French-speaking participants.** (A) A dendrogram representing the hierarchical clustering of language comprehension abilities. (B) Principal component analysis (PCA) plot, where ovals highlight clusters identified by UHCA. The PCA reveals a distinct separation among Command, Modifier and Syntactic Mechanisms. Principal component 1 accounts for 42.6% of the variance in the data. Principal component 2 accounts for 11.3% of the variance in the data.

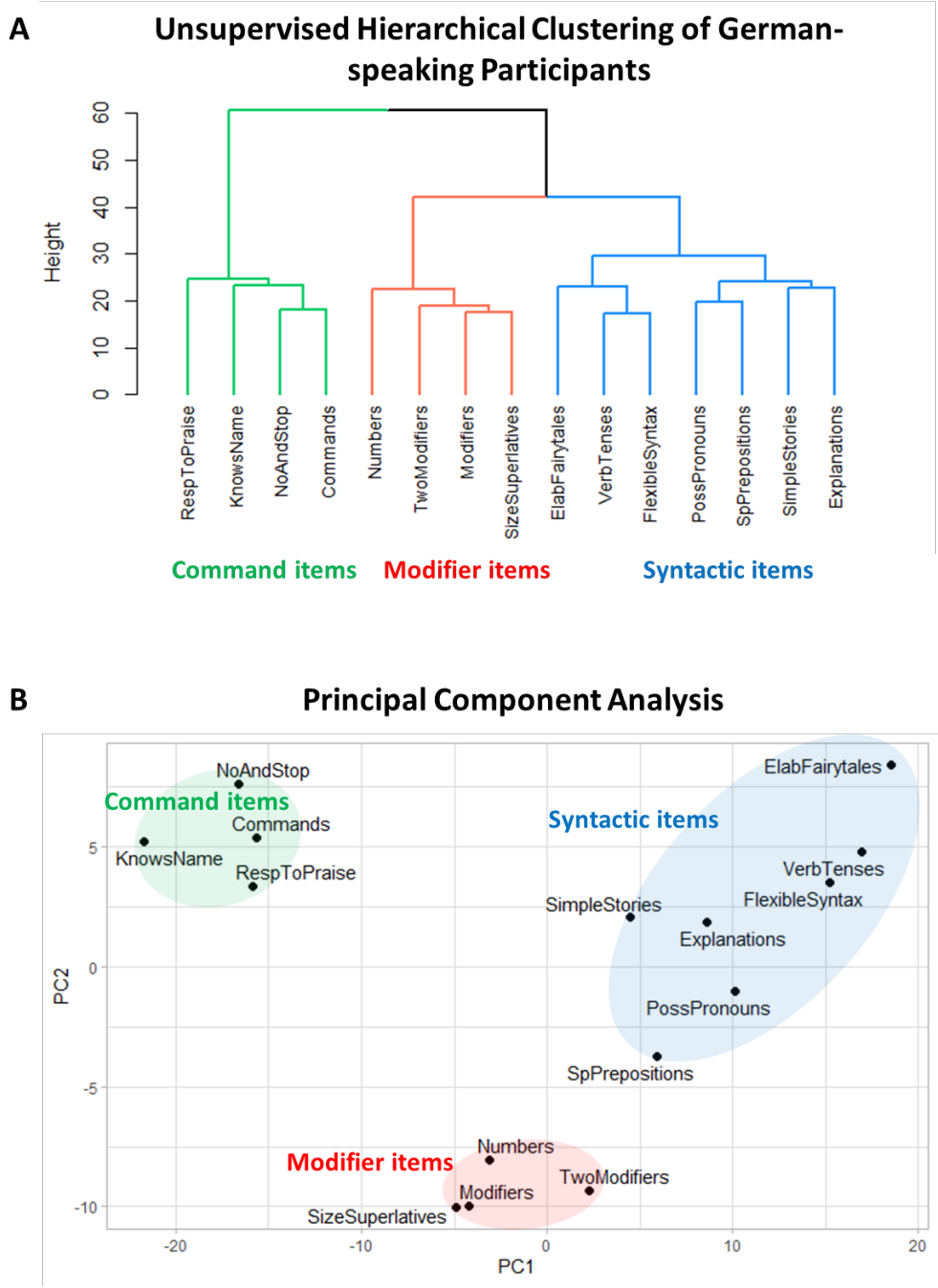

**Figure 8. Clustering analysis of 15 language comprehension items in German-speaking participants.** (A) A dendrogram representing the hierarchical clustering of language comprehension abilities. (B) Principal component analysis (PCA) plot, where ovals highlight clusters identified by UHCA. The PCA reveals a distinct separation among Command, Modifier

and Syntactic Mechanisms. Principal component 1 accounts for 42.4% of the variance in the data. Principal component 2 accounts for 10.7% of the variance in the data.

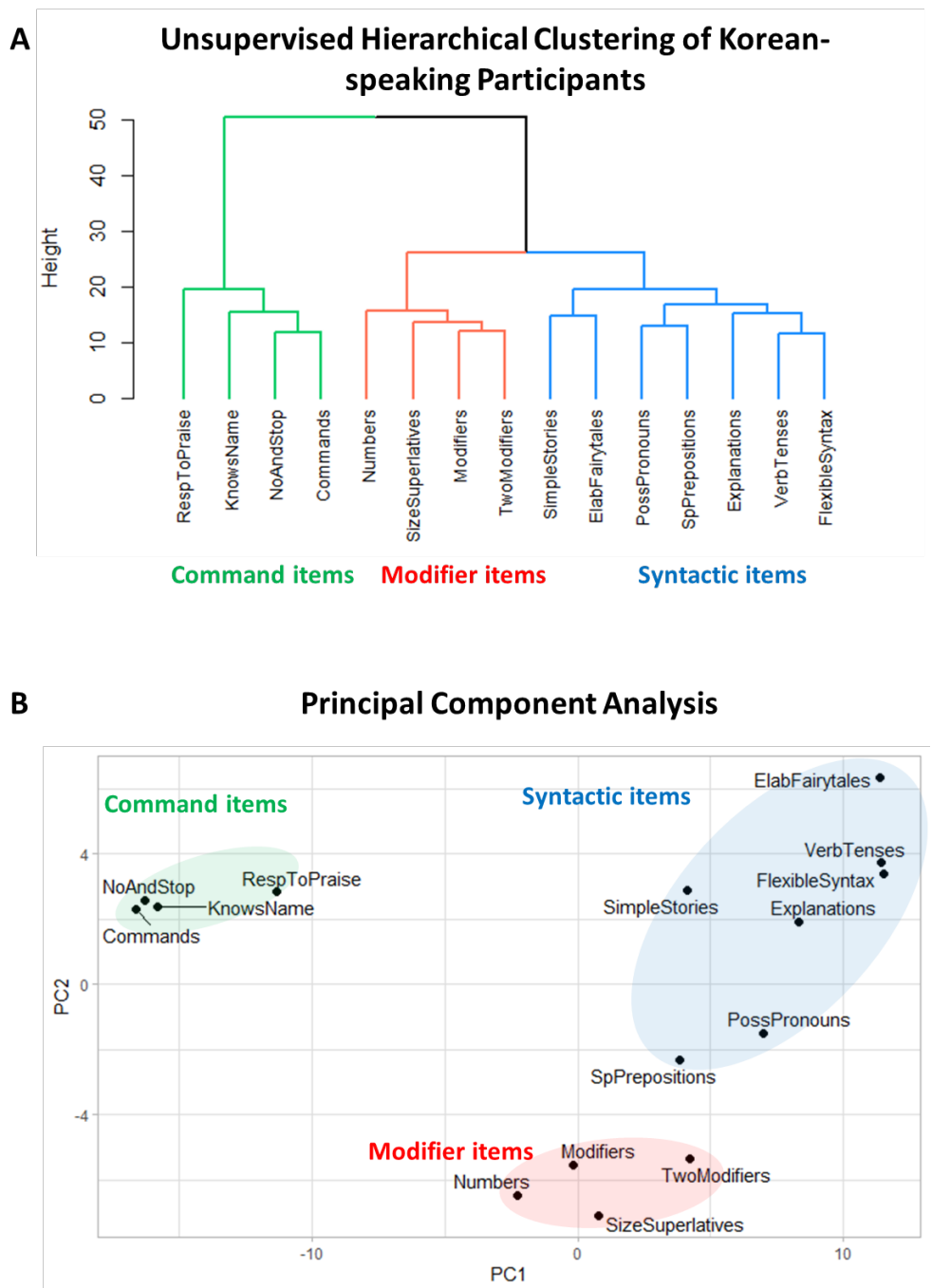

**Figure 9. Clustering analysis of 15 language comprehension items in Korean-speaking participants.** (A) A dendrogram representing the hierarchical clustering of language comprehension abilities. (B) Principal component analysis (PCA) plot, where ovals highlight clusters identified by UHCA. The PCA reveals a distinct separation among Command, Modifier

and Syntactic Mechanisms. Principal component 1 accounts for 49.5% of the variance in the data. Principal component 2 accounts for 8.7% of the variance in the data.

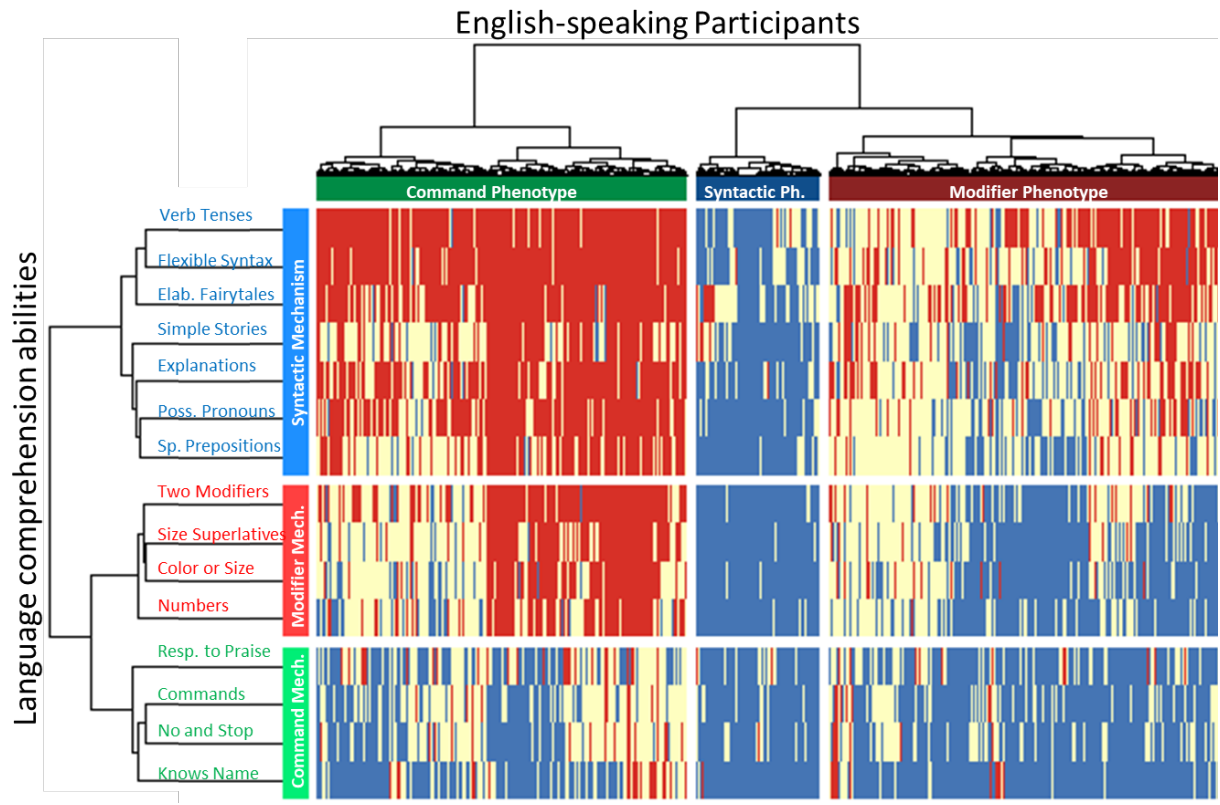

**Figure 10. Two-dimensional heatmap relating English-speaking participants to their language comprehension abilities.** The 15 language comprehension abilities are shown as rows. The dendrogram representing language comprehension abilities is shown on the left. Participants are shown as 27,187 columns. The dendrogram representing participants is shown on the top. Blue color indicates the presence of a linguistic ability (the "very true" answer), red indicates the lack of a linguistic ability (the "not true" answer), and white-yellow indicates the "somewhat true" answer.

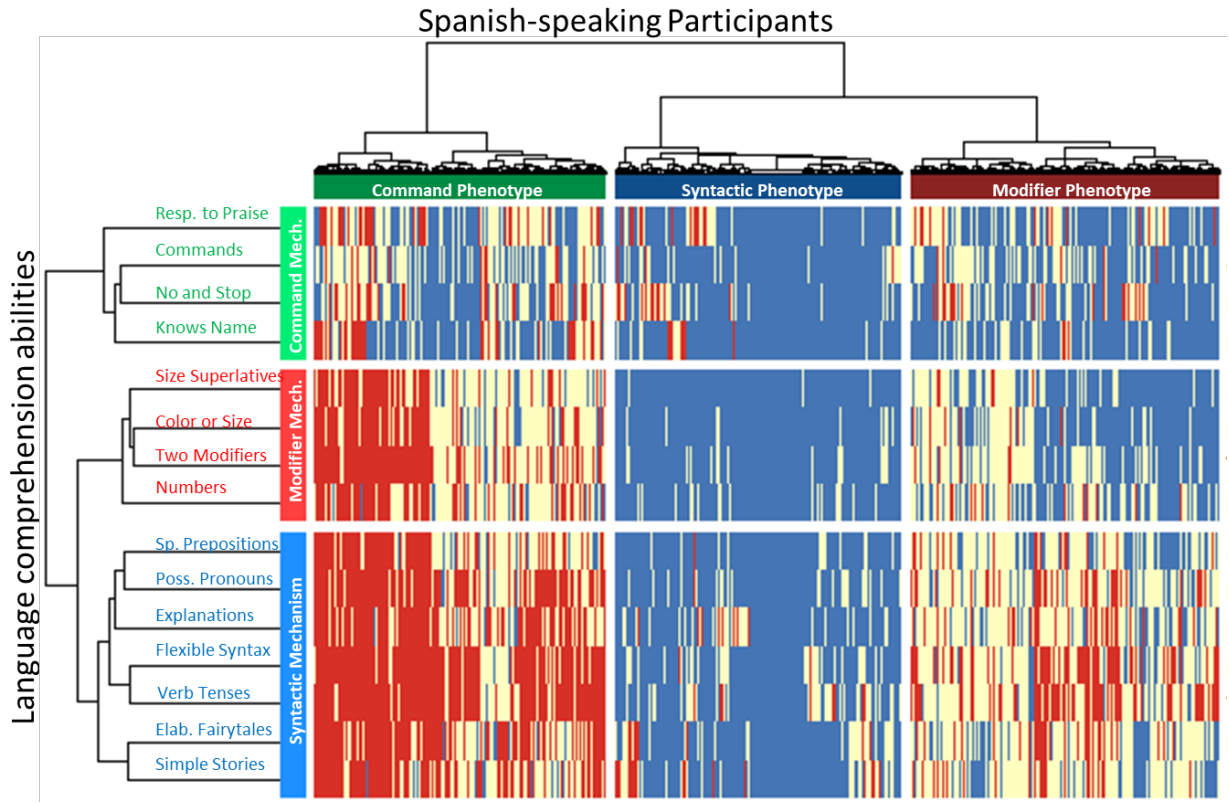

**Figure 11. Two-dimensional heatmap relating Spanish-speaking participants to their language comprehension abilities.** The 15 language comprehension abilities are shown as rows. The dendrogram representing language comprehension abilities is shown on the left. Participants are shown as 33,488 columns. The dendrogram representing participants is shown on the top. Blue color indicates the presence of a linguistic ability (the "very true" answer), red indicates the lack of a linguistic ability (the "not true" answer), and white-yellow indicates the "somewhat true" answer.

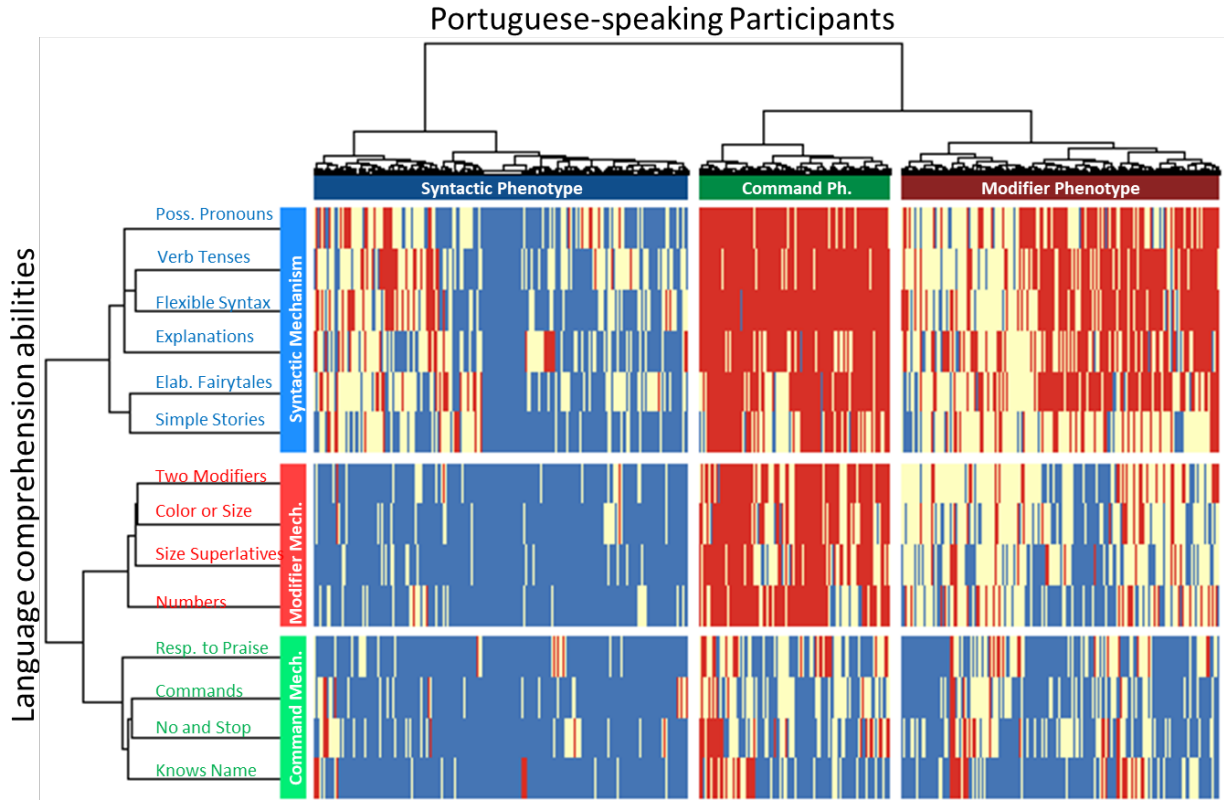

**Figure 12. Two-dimensional heatmap relating Portuguese-speaking participants to their language comprehension abilities.** The 14 language comprehension abilities are shown as rows. The dendrogram representing language comprehension abilities is shown on the left. Participants are shown as 7,504 columns. The dendrogram representing participants is shown on the top. Blue color indicates the presence of a linguistic ability (the "very true" answer), red indicates the lack of a linguistic ability (the "not true" answer), and white-yellow indicates the "somewhat true" answer.

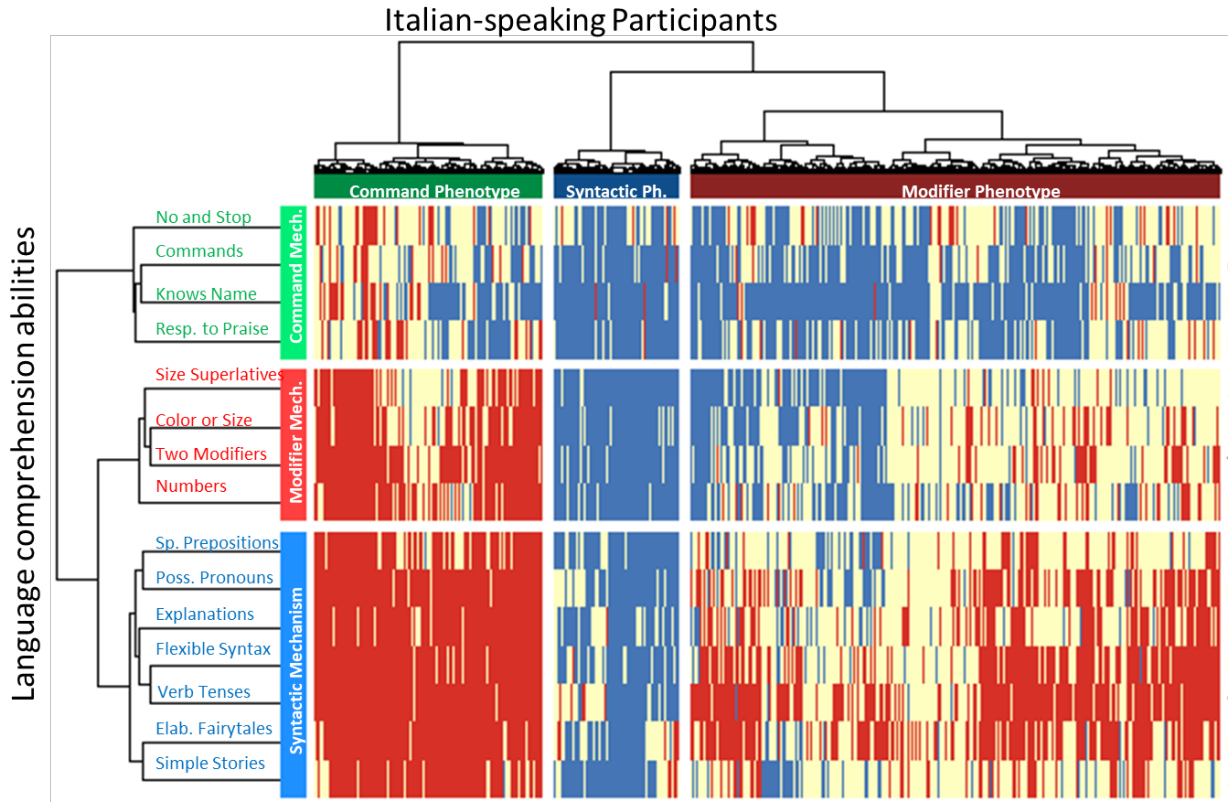

**Figure 13. Two-dimensional heatmap relating Italian-speaking participants to their language comprehension abilities.** The 15 language comprehension abilities are shown as rows. The dendrogram representing language comprehension abilities is shown on the left. Participants are shown as 6,484 columns. The dendrogram representing participants is shown on the top. Blue color indicates the presence of a linguistic ability (the "very true" answer), red indicates the lack of a linguistic ability (the "not true" answer), and white-yellow indicates the "somewhat true" answer.

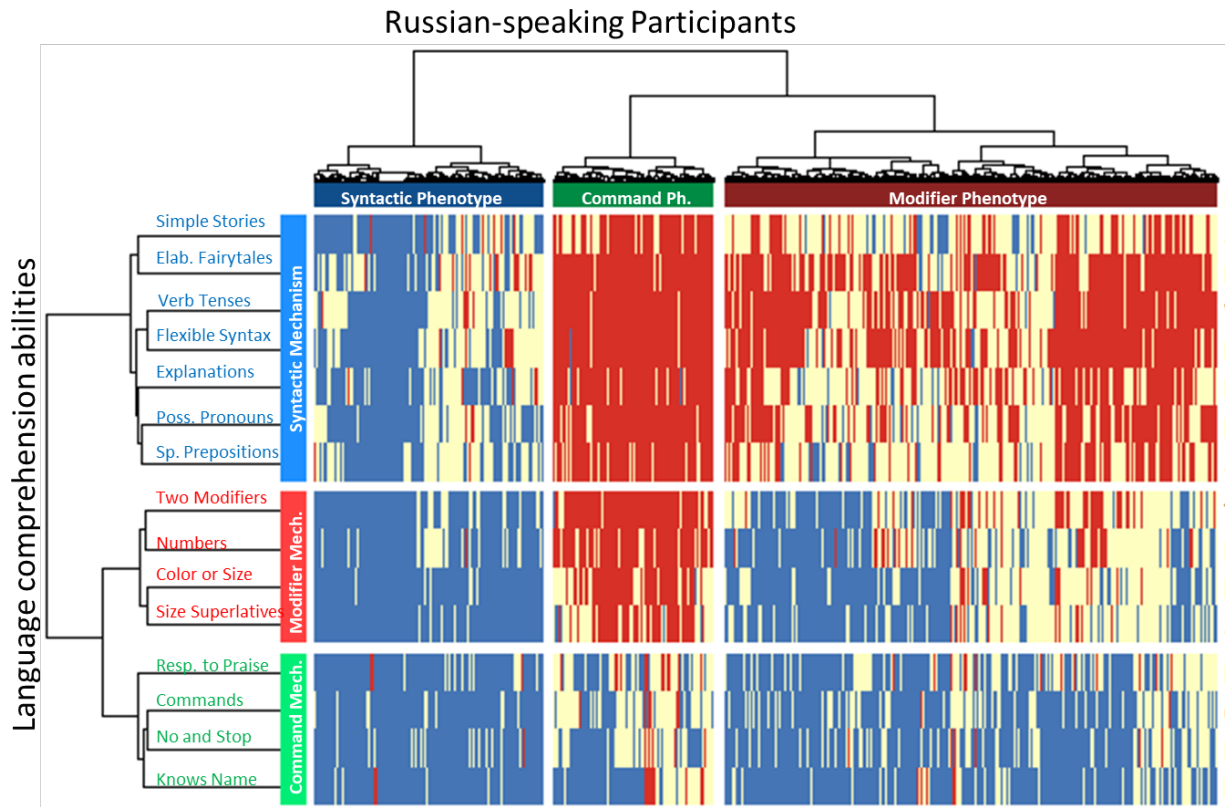

**Figure 14. Two-dimensional heatmap relating Russian-speaking participants to their language comprehension abilities.** The 15 language comprehension abilities are shown as rows. The dendrogram representing language comprehension abilities is shown on the left. Participants are shown as 4,778 columns. The dendrogram representing participants is shown on the top. Blue color indicates the presence of a linguistic ability (the "very true" answer), red indicates the lack of a linguistic ability (the "not true" answer), and white-yellow indicates the "somewhat true" answer.

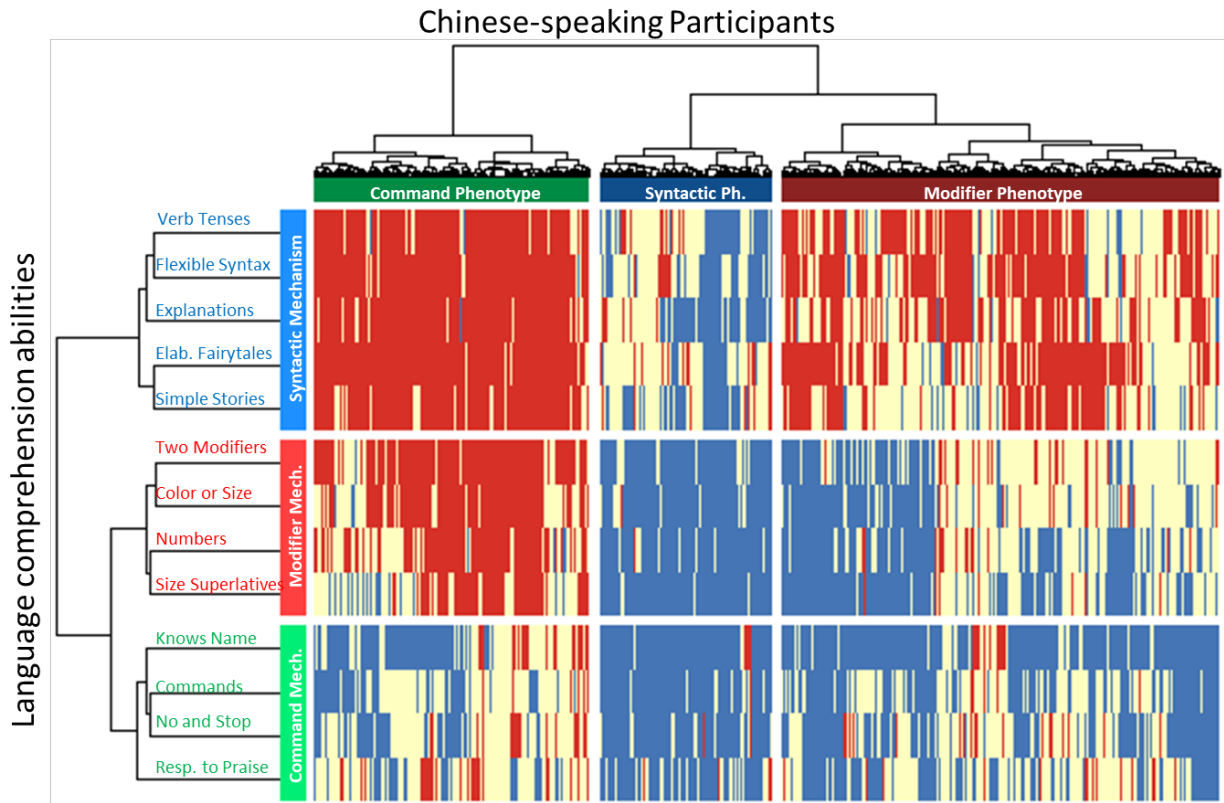

**Figure 15. Two-dimensional heatmap relating Chinese-speaking participants to their language comprehension abilities.** The 13 language comprehension abilities are shown as rows. The dendrogram representing language comprehension abilities is shown on the left. Participants are shown as 2,217 columns. The dendrogram representing participants is shown on the top. Blue color indicates the presence of a linguistic ability (the "very true" answer), red indicates the lack of a linguistic ability (the "not true" answer), and white-yellow indicates the "somewhat true" answer.

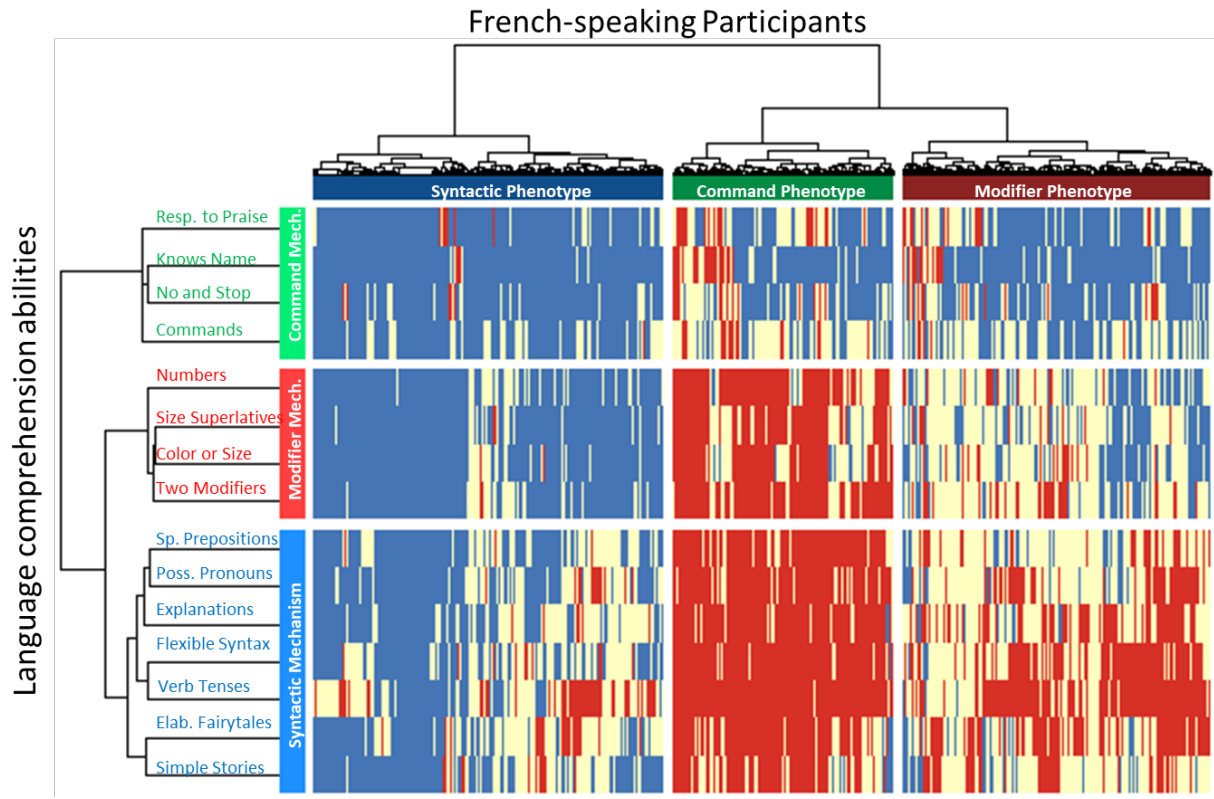

**Figure 16. Two-dimensional heatmap relating French-speaking participants to their language comprehension abilities.** The 15 language comprehension abilities are shown as rows. The dendrogram representing language comprehension abilities is shown on the left. Participants are shown as 1,060 columns. The dendrogram representing participants is shown on the top. Blue color indicates the presence of a linguistic ability (the "very true" answer), red indicates the lack of a linguistic ability (the "not true" answer), and white-yellow indicates the "somewhat true" answer.

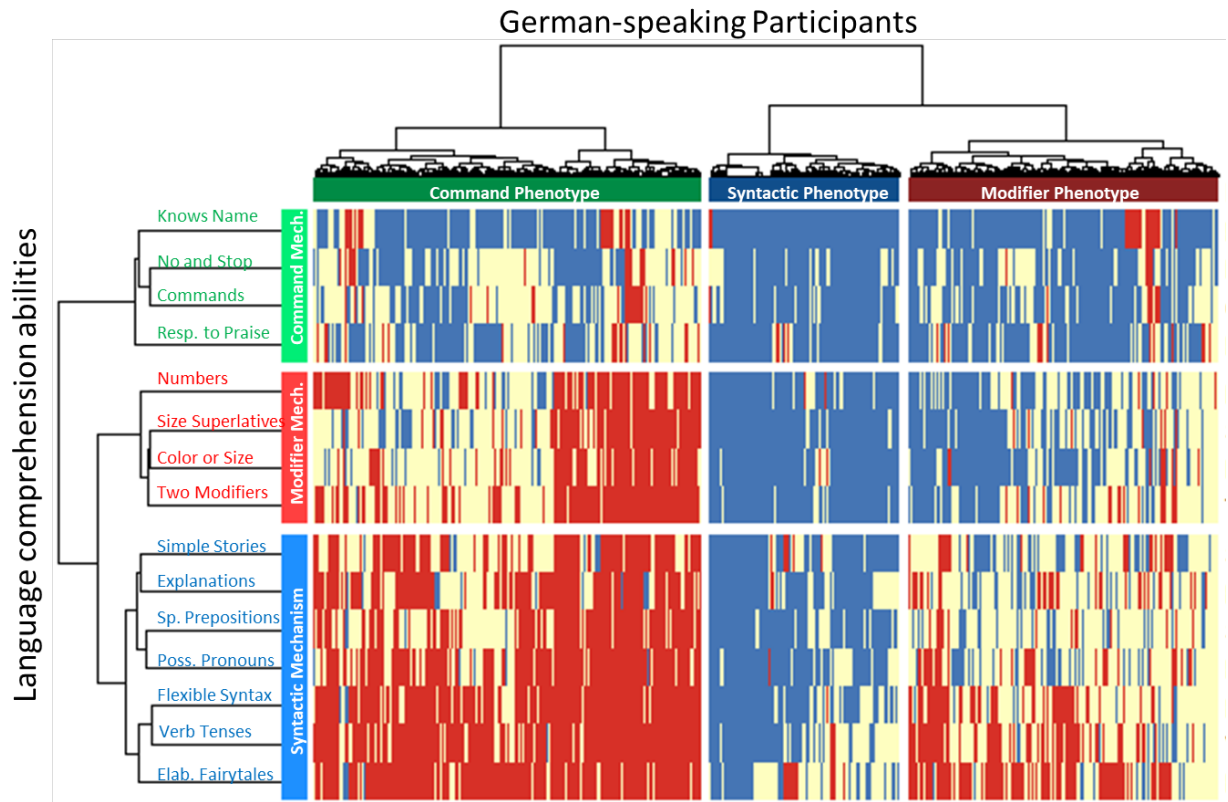

**Figure 17. Two-dimensional heatmap relating German-speaking participants to their language comprehension abilities.** The 15 language comprehension abilities are shown as rows. The dendrogram representing language comprehension abilities is shown on the left. Participants are shown as 927 columns. The dendrogram representing participants is shown on the top. Blue color indicates the presence of a linguistic ability (the "very true" answer), red indicates the lack of a linguistic ability (the "not true" answer), and white-yellow indicates the "somewhat true" answer.

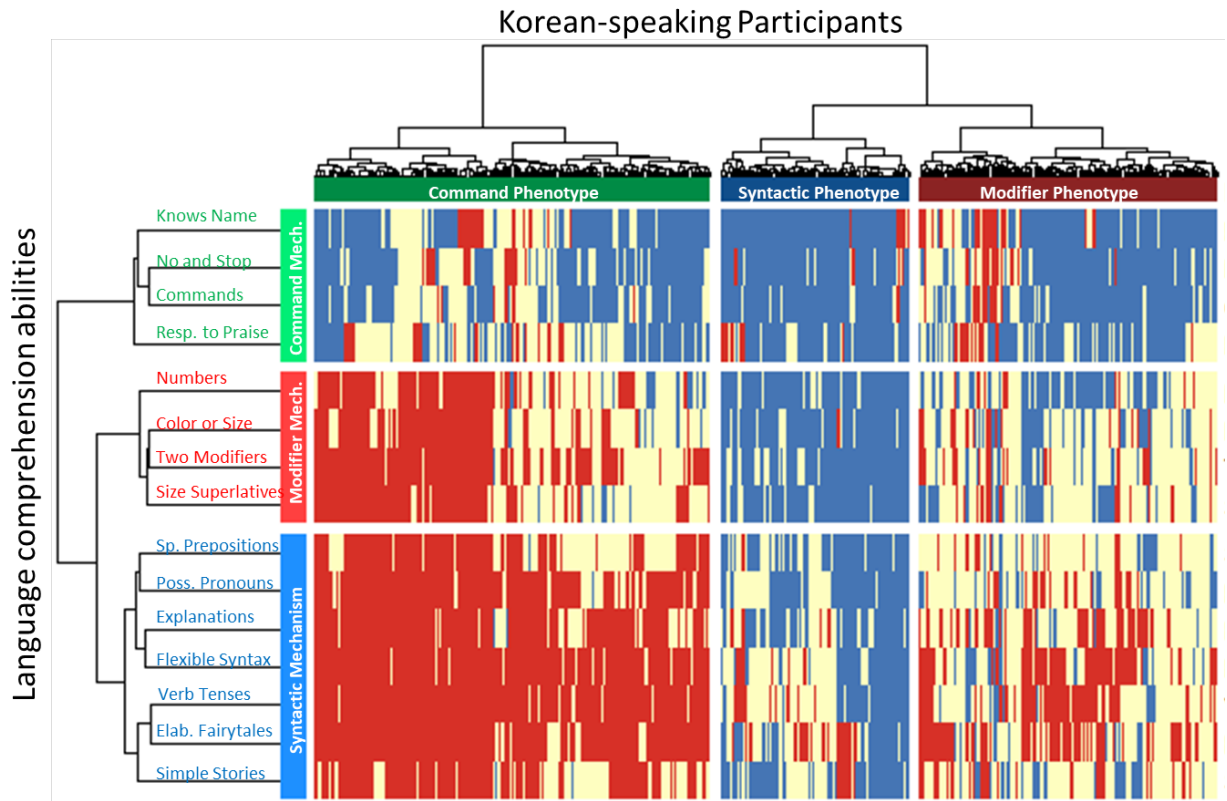

**Figure 18. Two-dimensional heatmap relating Korean-speaking participants to their language comprehension abilities.** The 15 language comprehension abilities are shown as rows. The dendrogram representing language comprehension abilities is shown on the left. Participants are shown as 454 columns. The dendrogram representing participants is shown on the top. Blue color indicates the presence of a linguistic ability (the "very true" answer), red indicates the lack of a linguistic ability (the "not true" answer), and white-yellow indicates the "somewhat true" answer.